# A Comparative Study between Moravec and Harris Corner Detection of Noisy Images Using Adaptive Wavelet Thresholding Technique

**Nilanjan Dey[1], Pradipti Nandi [2] , Nilanjana Barman[3] , Debolina Das[4] , Subhabrata Chakraborty[5]**

[1]Asst. Professor, Dept. of IT, JIS College of Engineering, Kalyani, West Bengal, India.
[2, 3,4,5]B Tech Student,Dept. of CSE, JIS College of Engineering, Kalyani, West Bengal, India.

## ABSTRACT

**In this paper a comparative study between Moravec and Harris Corner Detection has been done for obtaining features required to track and recognize objects within a noisy image. Corner detection of noisy images is a challenging task in image processing. Natural images often get corrupted by noise during acquisition and transmission. As Corner detection of these noisy images does not provide desired results, hence de-noising is required. Adaptive wavelet thresholding approach is applied for the same.**

## I. Introduction

A corner is a point for which there are two dominant and different edge directions in the vicinity of the point. In simpler terms, a corner can be defined as the intersection of two edges, where an edge is a sharp change in image brightness. Generally termed as interest point detection, corner detection is a methodology used within computer vision systems to obtain certain kinds of features from a given image. The initial operator concept of "points of interest" in an image, which could be used to locate matching regions in different images, was developed by Hans P. Moravec in 1977. The Moravec operator is considered to be a corner detector because it defines interest points as points where there are large intensity variations in all directions.

For a human, it is easier to identify a “corner”, but a mathematical detection is required in case of algorithms. Chris Harris and Mike Stephens in 1988 improved upon Moravec's corner detector by taking into account the differential of the corner score with respect to direction directly, instead of using shifted patches. Moravec only considered shifts in discrete 45 degree angles whereas Harris considered all directions. Harris detector has proved to be more accurate in distinguishing between edges and corners. He used a circular Gaussian window to reduce noise. Still in cases of noisy images, it’s difficult to find out the exact number of corners. One of the most conventional ways of image de-noising is using linear filters like Wiener filter. In the presence of additive noise the resultant noisy image, through linear filters, gets blurred and smoothed with poor feature localization and incomplete noise suppression. To overcome these limitations, nonlinear filters have been proposed like adaptive wavelet thresholding approach.

Adaptive wavelet thresholding approach gives a very good result for the same. Wavelet Transformation has its own excellent space-frequency localization property and thresholding removes coefficients that are inconsiderably relative to some adaptive data-driven threshold.

## II. Discrete wavelet transformation

The wavelet transform describes a multi-resolution decomposition process in terms of expansion of an image onto a set of wavelet basis functions. Discrete Wavelet Transformation has its own excellent space frequency localization property. Applying DWT in 2D images corresponds to 2D filter image processing in each dimension. The input image is divided into 4 non-overlapping multi-resolution sub-bands by the filters, namely LL1 (Approximation coefficients), LH1 (vertical details), HL1 (horizontal details) and HH1 (diagonal details). The sub-band (LL1) is processed further to obtain the next coarser scale of wavelet coefficients, until some final scale “N” is reached. When “N” is reached, we’ll have 3N+1 sub-bands consisting of the multi-resolution sub-bands (LLN) and (LHX), (HLX) and (HHX) where “X” ranges from 1 until “N”. Generally most of the Image energy is stored in these sub-bands.

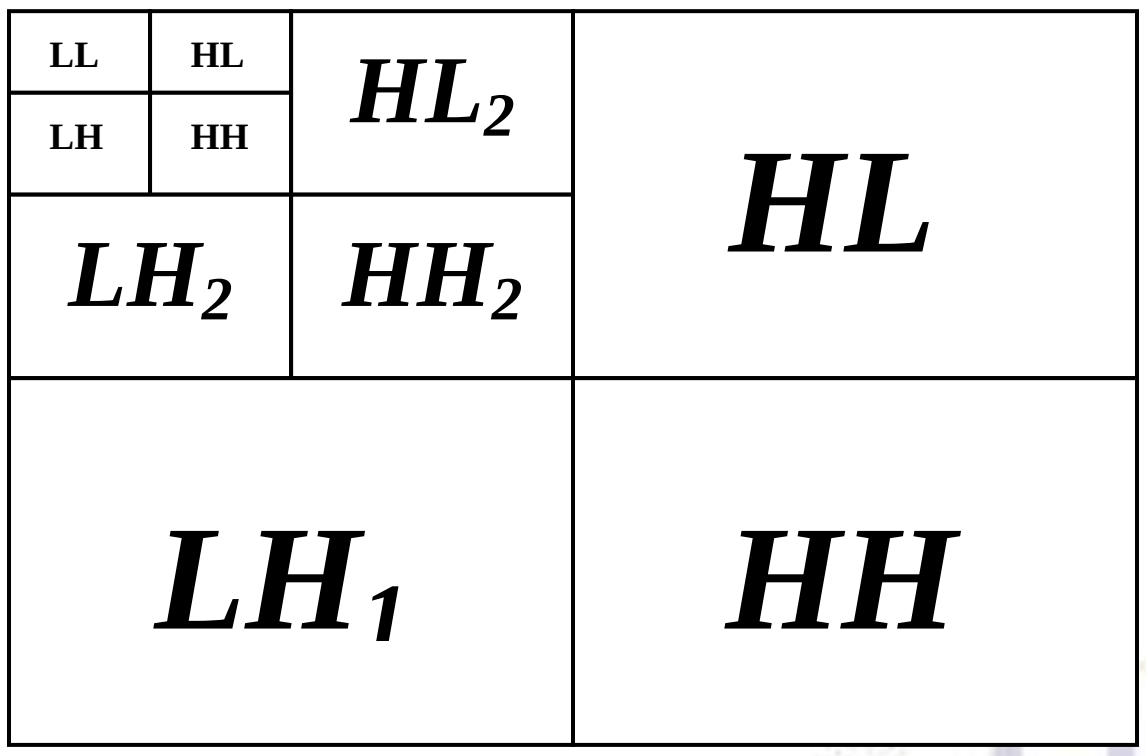

Figure 1. Three phase decomposition using DWT.

The Haar wavelet is also the simplest possible wavelet. Haar wavelet is not continuous, and therefore not differentiable. This property can, however, be an advantage for the analysis of signals with sudden transitions.

## III. Wavelet Thresholding

The concept of wavelet de-noising technique can be given as follows. Assuming that the noisy data is given by the following equation,

X (t) = S (t) + N (t)

Where, S (t) is the uncorrupted signal with additive noise N (t). Let W (.) and $W^{-1}$(.) denote the forward and inverse wavelet transform operators.

Let D (., λ) denote the de-noising operator with threshold λ. We intend to de-noise X (t) to recover Ŝ (t) as an estimate of S (t).

The technique can be summarized in three steps

Y = W(X) ..... (2)

Z = D(Y, λ) ..... (3)

Ŝ = $W^{-1}$ (Z) ..... (4)

D (., λ) being the thresholding operator and λ being the threshold.

A signal estimation technique that exploits the potential of wavelet transform required for signal de-noising is called Wavelet Thresholding [1, 2, 3]. It de-noises by eradicating coefficients that are extraneous relative to some threshold.

There are two types of recurrently used thresholding methods, namely hard and soft thresholding [4, 5].

The Hard thresholding method zeros the coefficients that are smaller than the threshold and leaves the other ones unchanged. On the other hand soft thresholding scales the remaining coefficients in order to form a continuous distribution of the coefficients centered on zero.

The hard thresholding operator is defined as

D (U, λ) = U for all |U|> λ

Hard threshold is a keep or kill procedure and is more intuitively appealing. The hard-thresholding function chooses all wavelet coefficients that are greater than the given λ (threshold) and sets the other to zero. λ is chosen according to the signal energy and the noise variance ($\sigma^2$)

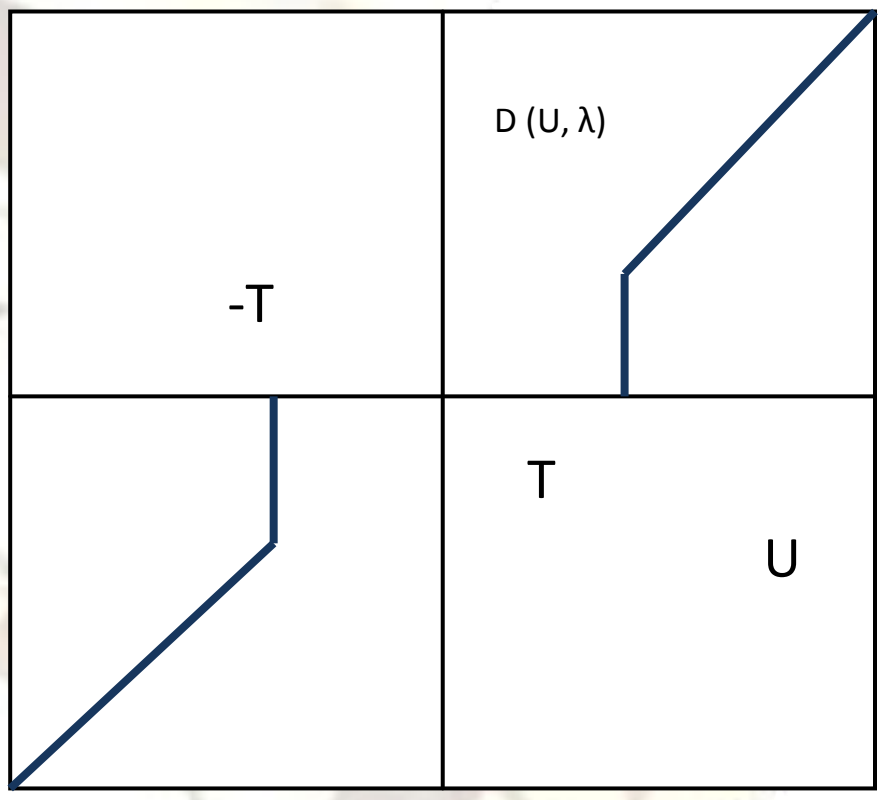

Figure 2. Hard Thresholding

The soft thresholding operator is defined as

D (U, λ) = sgn (U) max (0, |U| - λ)

Soft thresholding shrinks wavelets coefficients by λ towards zero.

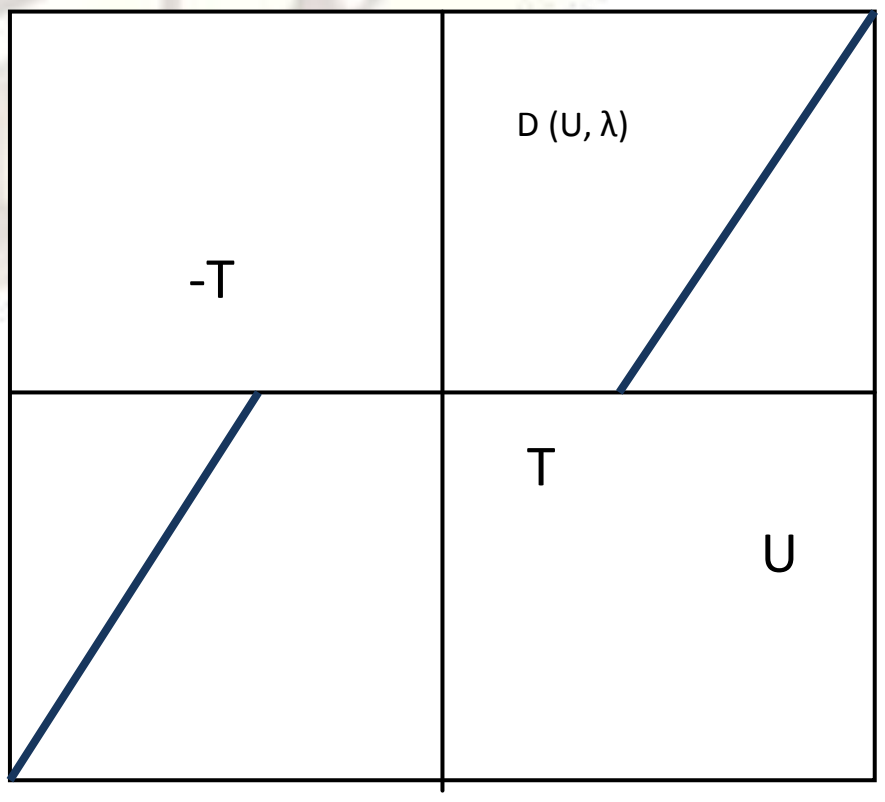

Figure 3. Soft Thresholding

## IV. Bayes Shrink (BS)

Bayes Shrink, [6, 7] proposed by Chang Yu and Vetterli, is an adaptive data-driven threshold for image de-noising via wavelet soft-thresholding. Generalized Gaussian distribution (GGD) for the wavelet coefficients is assumed in each detail sub band. It is then tried to estimate the threshold T which minimizes the Bayesian Risk, which gives the name Bayes Shrink.

It uses soft thresholding which is done at each band of resolution in the wavelet decomposition. The Bayes threshold, $T_B$, is defined as

$$T_B = \sigma^2 / \sigma_s \qquad \ldots\ldots\ldots\ldots\ldots\ldots\ldots (5)$$

Where $\sigma^2$ is the noise variance and $\sigma_s^2$ is the signal variance without noise. The noise variance $\sigma^2$ is estimated from the sub band $HH_1$ by the median estimator

$$\hat{\sigma} = \frac{median\left(\left\{\left| g_{j-1,k} \right| : k = 0,1,\ldots,2^{j-1}-1\right\}\right)}{0.6745} \quad ..(6)$$

where $g_{j-1,k}$ corresponds to the detail coefficients in the wavelet transform. From the definition of additive noise we have

$$w(x, y) = s(x, y) + n(x, y)$$

Since the noise and the signal are independent of each other, it can be stated that

$$\sigma_w^2 = \sigma_s^2 + \sigma^2$$

$\sigma_w^2$ can be computed as shown :

$$\sigma^2{}_w = \frac{1}{n^2}\sum_{x,y=1}^{n} w^2(x,y)$$

The variance of the signal, $\sigma_s^2$ is computed as

$$\sigma_s = \sqrt{\max(\sigma^2{}_w - \sigma^2, 0)}\,.$$

With $\sigma_2$ and $\sigma^2_s$, the Bayes threshold is computed from Equation (5).

## V. Moravec Corner Detection

Hans P. Moravec developed Moravec operator in 1977 for his research involving the navigation of the Stanford Cart [10,11] through a clustered environment. Since it defines interest points as points where there is a large intensity variation in every direction, which is the case at corners, the Moravec operator is considered a corner detector.

However, Moravec was not specifically interested in finding corners, just distinct regions in an image that could be used to register consecutive image frames.

The concept of "points of interest" as distinct regions in images was defined by him. It was concluded that in order to find matching regions in consecutive image frames, these interest points could be used. In determining the existence and location of objects in the vehicle's environment, this proved to be a vital low-level processing step.

Since the concept of a corner is not well-defined for gray scale images, many have commended this relaxation in the "definition" of “a corner”.

***Algorithm***

The Moravec corner detector is stated formally below:[12]
Denote the image intensity of a pixel at (x, y) by I(x, y).
Input: grayscale image, window size, threshold T
Output: map indicating position of each detected corner.

1. For each pixel (x, y) in the image calculate the intensity variation from a shift (u, v) as:

$$V_{u,v}(x,y) = \sum_{\forall a,b \text{ in the window}} \left(I(x+u+a, y+v+b) - I(x+a, y+b)\right)^2$$

where the shifts (u,v) considered are: (1,0),(1,1),(0,1),(-1,1),(-1,0),(-1,-1),(0,-1),(1,-1)

2. Construct the cornerness map by calculating the cornerness measure C(x, y) for each pixel (x, y):

$$C(x,y) = \min\left(V_{u,v}(x,y)\right)$$

3. Threshold the interest map by setting all C(x, y) below a threshold T to zero.

4. Perform non-maximal suppression to find local maxima.

All non-zero points remaining in the cornerness map are corners.

## VI. Harris Corner Detection

Harris corner detector [8,9] is based on the local auto-correlation function of a signal which measures the local changes of the signal with patches shifted by a small amount in different directions. Given a shift (x, y) and a point the auto-correlation function is defined as

$$c(x,y) = \sum_W [I(x_i,y_i) - I(x_i + \Delta x, y_i + \Delta y)]^2 \quad \text{....... (7)}$$

Where I ($x_i$, $y_i$) represent the image function and ($x_i$, $y_i$) are the points in the window W centered on (x, y).
The shifted image is approximated by a Taylor expansion truncated to the first order terms

$$I(x_i + \Delta x, y_i + \Delta y) \approx \left[ I(x_i,y_i) + [I_x(x_i,y_i) \;\; I_y(x_i,y_i)] \right] \begin{bmatrix} \Delta x \\ \Delta y \end{bmatrix} \quad \text{........... (8)}$$

where $I_x$ ($x_i$, $y_i$) and $I_y$ ($x_i$, $y_i$) indicate the partial derivatives in x and y respectively. With a filter like [-1, 0, 1] and [-1, 0, 1] $^T$, the partial derivates can be calculated from the image by substituting (8) in (7).

$$c(x,y) = [\Delta x \;\; \Delta y] \begin{bmatrix} \sum_W (I_x(x_i,y_i))^2 & \sum_W I_x(x_i,y_i)\, I_y(x_i,y_i) \\ \sum_W I_x(x_i,y_i)\, I_y(x_i,y_i) & \sum_W (I_y(x_i,y_i))^2 \end{bmatrix} \begin{bmatrix} \Delta x \\ \Delta y \end{bmatrix} = [\Delta x \;\; \Delta y]\, C(x,y) \begin{bmatrix} \Delta x \\ \Delta y \end{bmatrix}$$

C(x, y) the auto-correlation matrix which captures the intensity structure of the local neighborhood.
Let $\alpha_1$ and $\alpha_2$ be the eigen values of C(x, y), then we have 3 cases to consider:

1. Both Eigen values are small means uniform region (constant intensity).
2. Both Eigen values are high means Interest point (corner)
3. One Eigen value is high means contour(edge)

To find out the interest points, Characterize corner response H(x, y) by Eigen values of C(x, y).

- C(x, y) is symmetric and positive definite that is $\alpha_1$ and $\alpha_2$ are >0
- $\alpha_1 \alpha_2$ = det (C(x, y)) = AC –$B^2$
- $\alpha_1 + \alpha_2$ = trace(C(x, y)) = A + C
- Harris suggested: That the $H_{cornerResponse} = \alpha_1 \alpha_2 - 0.04(\alpha_1 + \alpha_2)^2$

Finally, it is needed to find out corner points as local maxima of the corner response.

## VII. Proposed Method

Step 1. Perform 2-level Multi-wavelet decomposition of the image corrupted by Gaussian noise.
Step 2. Apply Bayes Soft thresholding to the noisy coefficients.
Step 3. Apply Moravec and Harris corner detection on the de-noised image respectively.

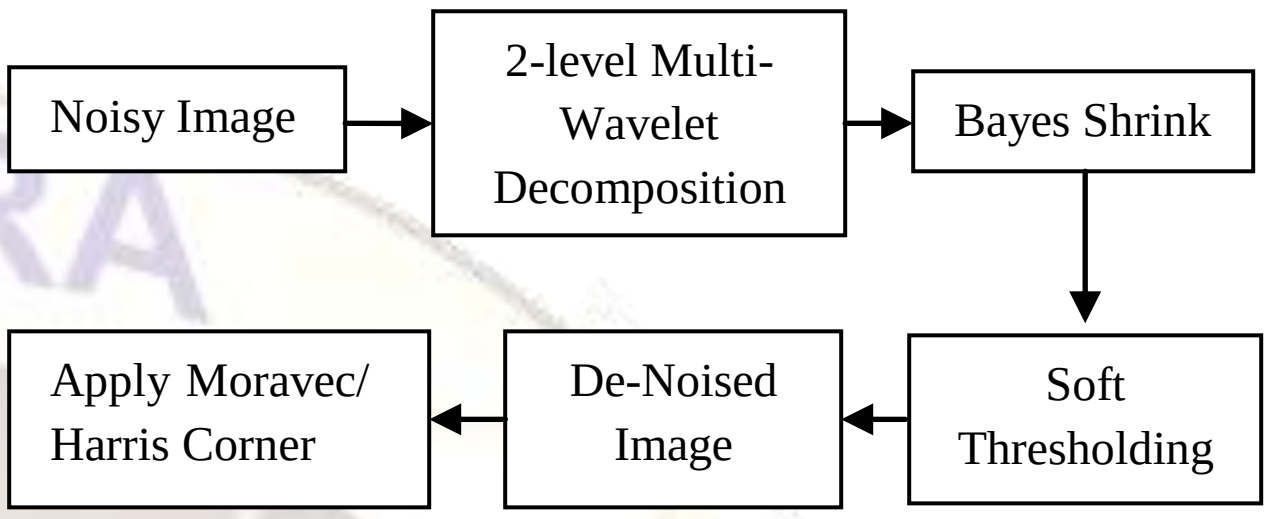

Figure 4. Corner detection of De-noised image

## VIII. Result and Discussions

Signal-to-noise ratio can be defined in a different manner in image processing where the numerator is the square of the peak value of the signal and the denominator equals the noise variance. Two of the error metrics used to compare the various image de-noising techniques is the Mean Square Error (MSE) and the Peak Signal to Noise Ratio (PSNR).

### Mean Square Error (MSE):

Mean Square Error is the measurement of average of the square of errors and is the cumulative squared error between the noisy and the original image.

$$MSE = \frac{1}{M \times N} \sum_{i=0}^{M-1} \sum_{j=0}^{N-1} (f(i,j) - g(i,j))^2$$

where f(i,j) and g(i,j) are the original secret image and extracted secret image with coordinate position (i,j).

### Peak Signal to Noise Ratio (PSNR):

PSNR is a measure of the peak error. Peak Signal to Noise Ratio is the ratio of the square of the peak value the signal could have to the noise variance.

$$PSNR = 10 * \log_{10} \left( \frac{255^2}{MSE} \right)$$

A higher value of PSNR is good because of the superiority of the signal to that of the noise.
MSE and PSNR values of an image are evaluated after adding Gaussian and Speckle noise. The following tabulation shows the comparative study based on Wavelet thresholding techniques of different decomposition levels.

Table 1

| **Noise Type** | **Wavelet** | **Thres-holding** | **Level of Decom-position** | **PSNR** |
|---|---|---|---|---|
| Gaussian | Haar | Bayes Soft | 1 | 22.8685 |
| | | | 2 | 23.6533 |
| Speckle | | | 1 | 23.0867 |
| | | | 2 | 23.6360 |
| Salt & Paper | | | 1 | 19.5202 |
| | | | 2 | 19.5405 |

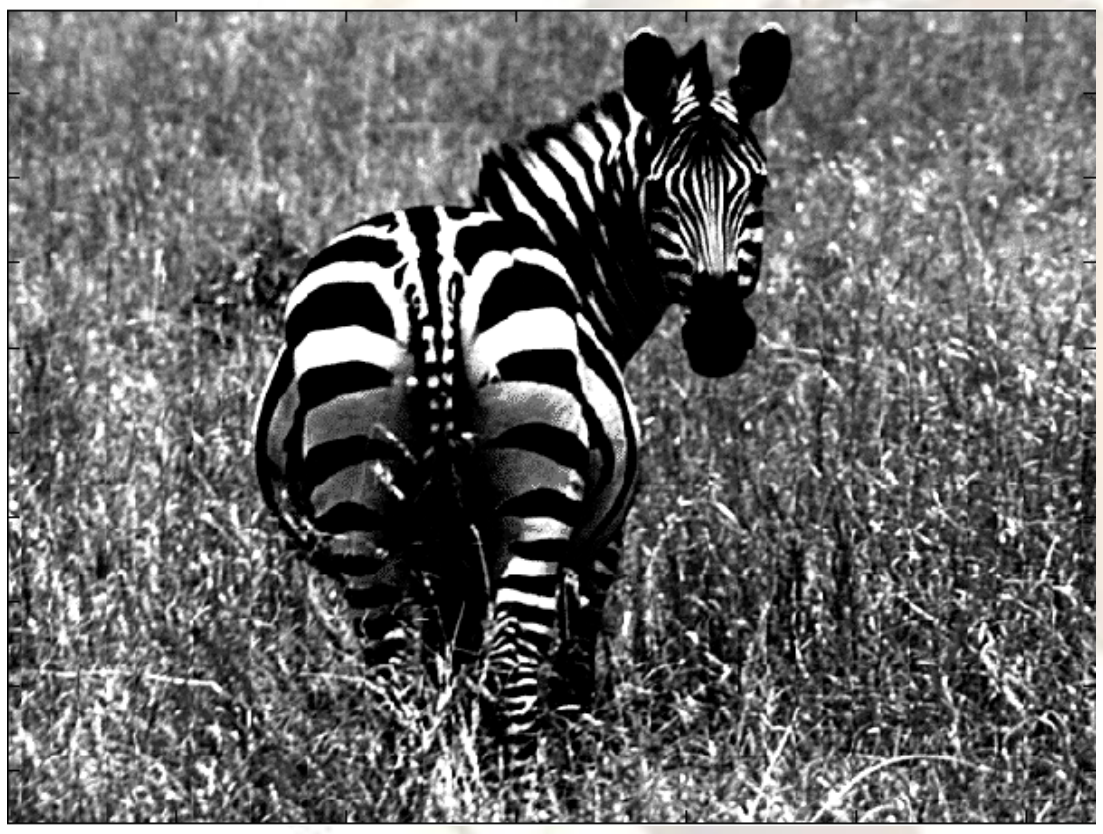

(a)

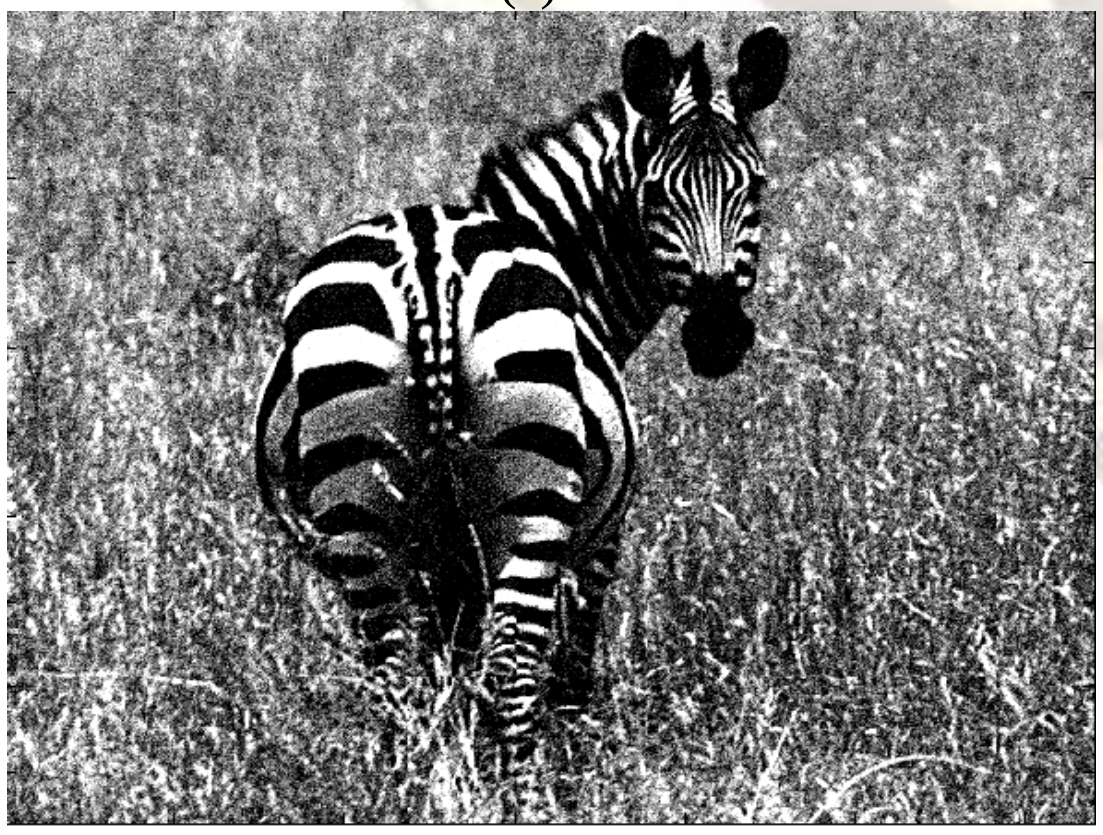

(b)

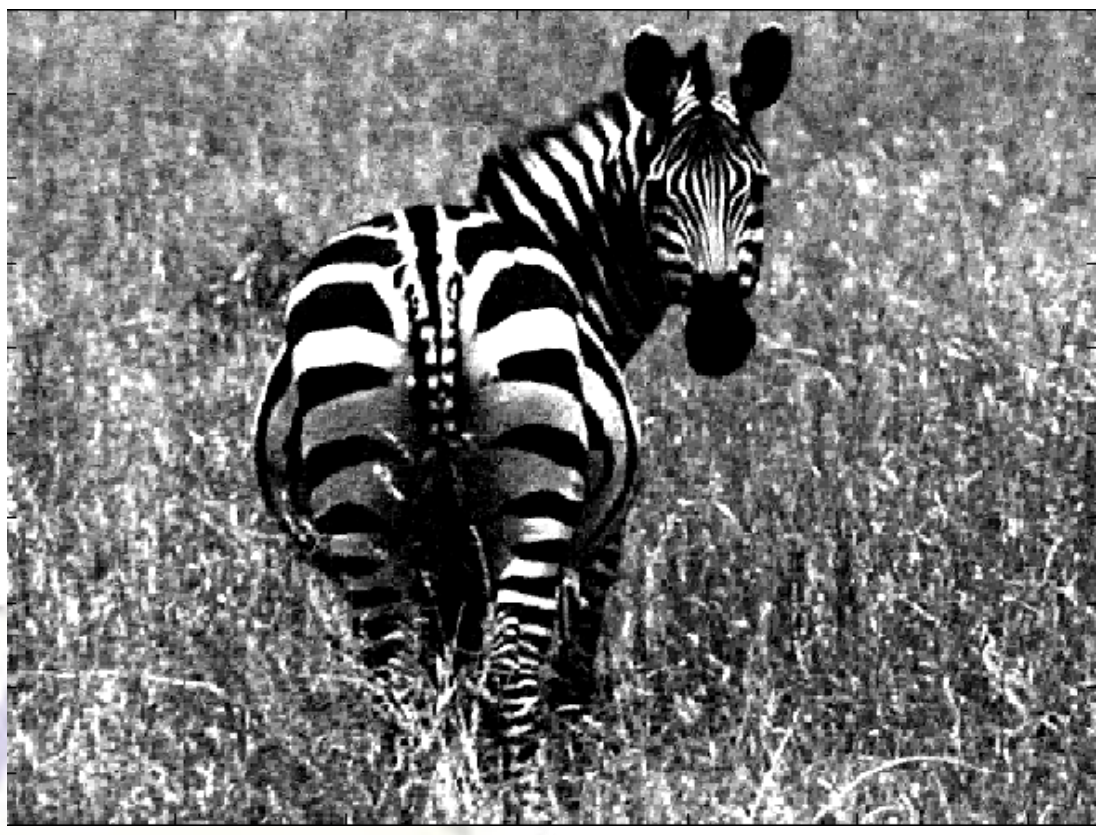

(c)

(a) Original Image (b) Noisy image (Gaussian) (c) Second level DWT decomposed and Bayes soft threshold noisy image

Figure 5. De-noised Image

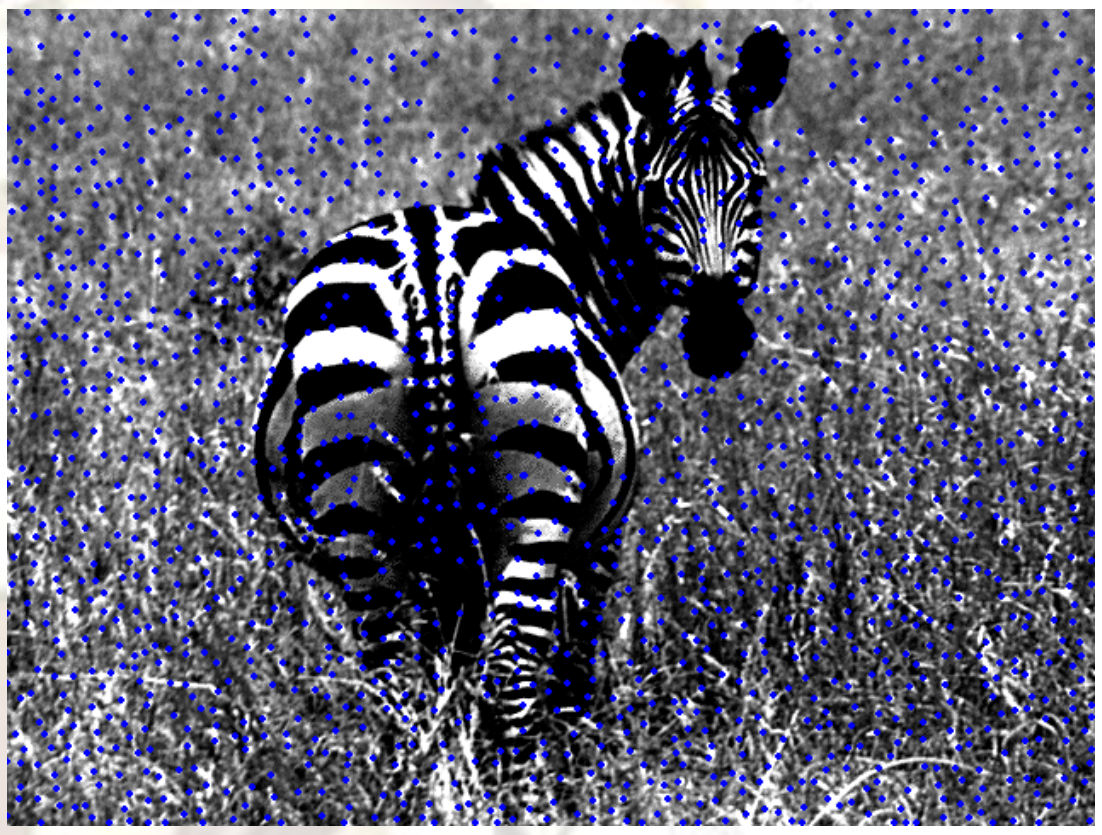

(d)

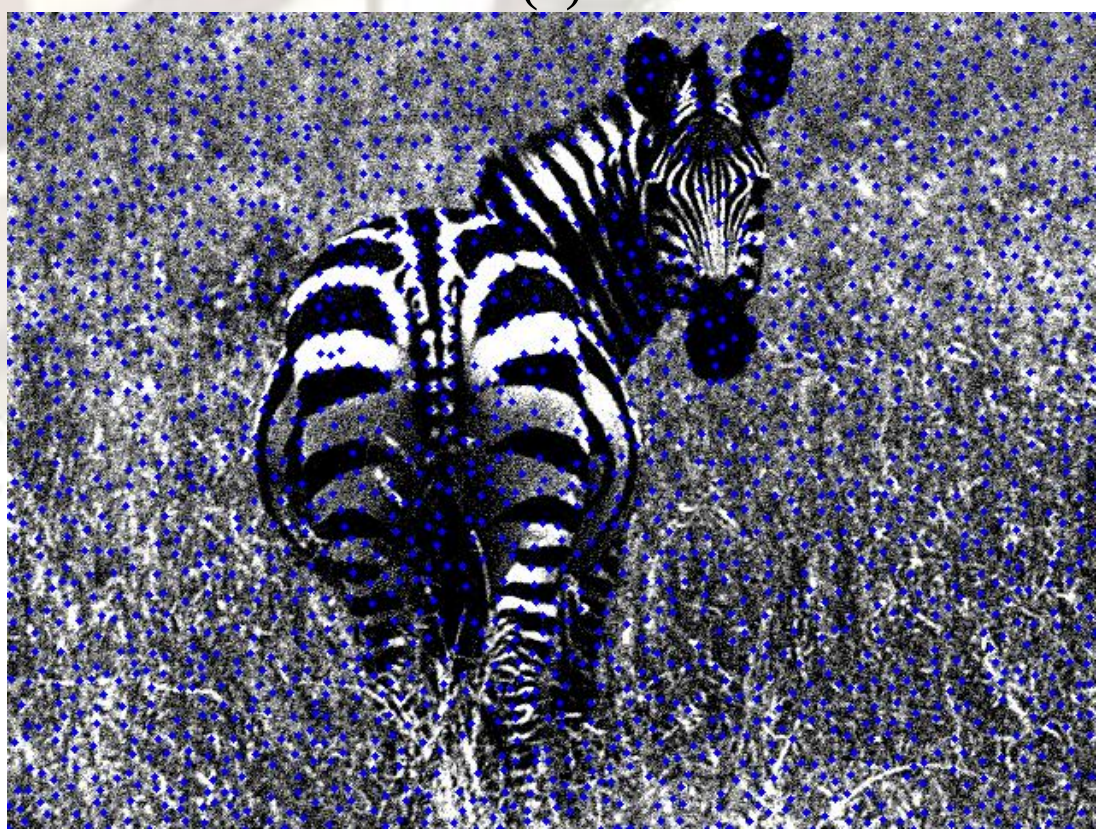

(e)

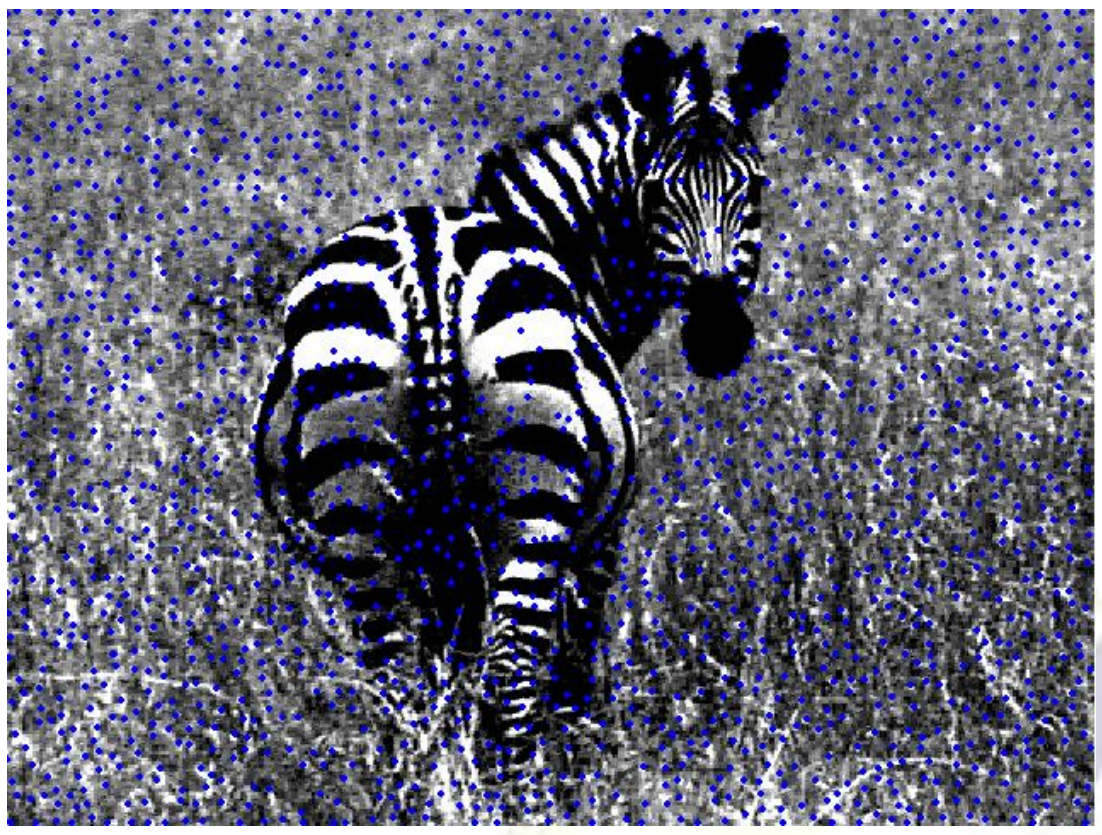

(f)

(d) Moravec Corner Detection on original image (e) Moravec Corner Detection on noise image (f) Moravec Corner Detection on de-noised image using Bayes soft threshold method.

Figure 6. Moravec Corner Detection

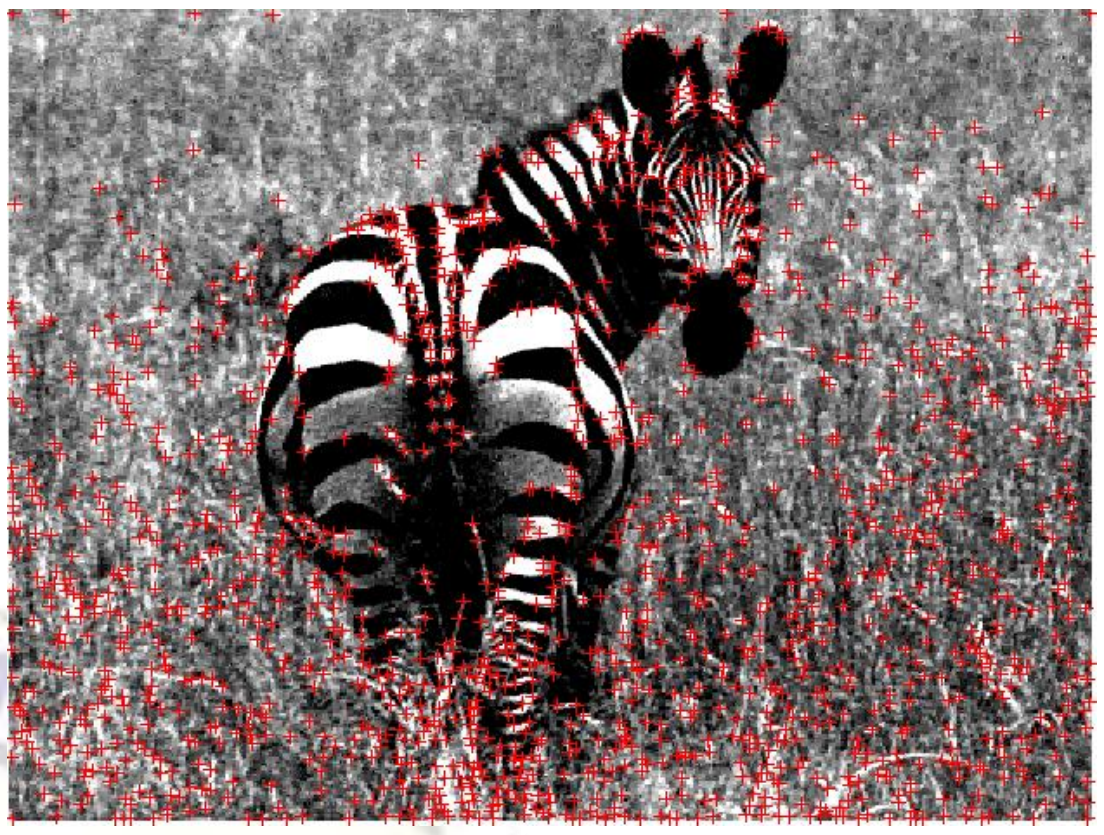

(i)

(g) Harris Corner Detection on original image (h) Harris Corner Detection on noise image (i) Harris Corner Detection on de-noised image using Bayes soft threshold method.

Figure 7. Harris Corner Detection

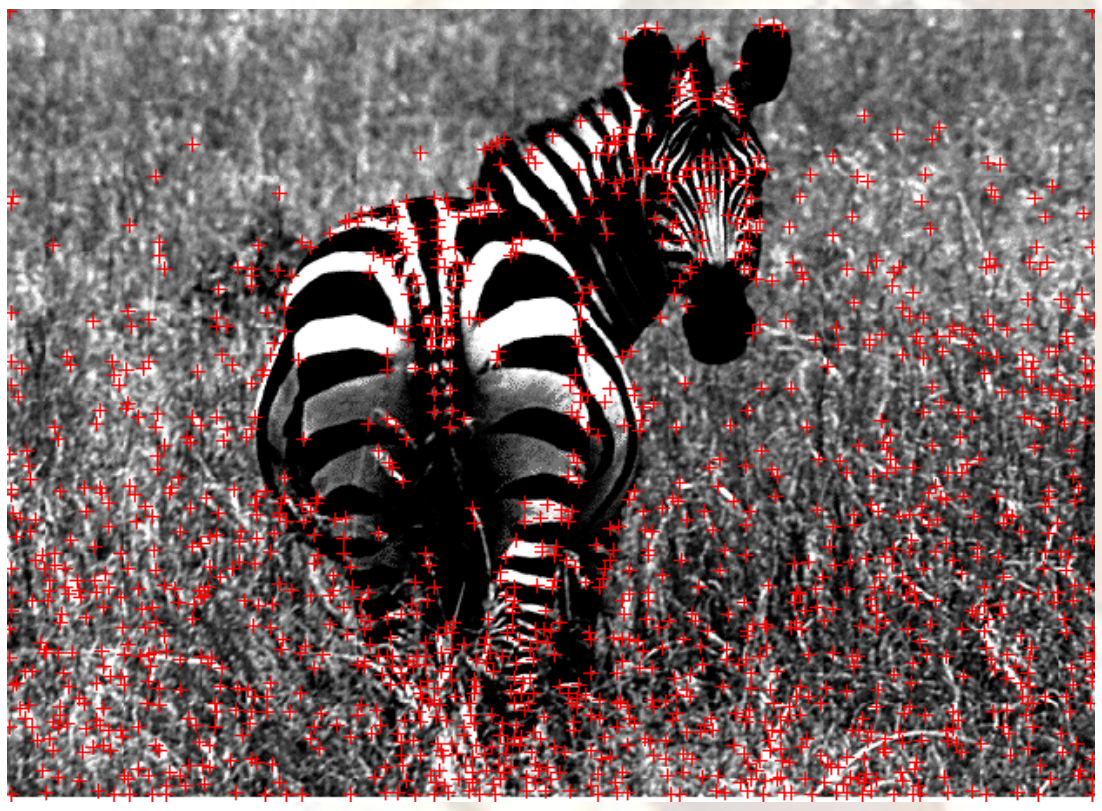

(g)

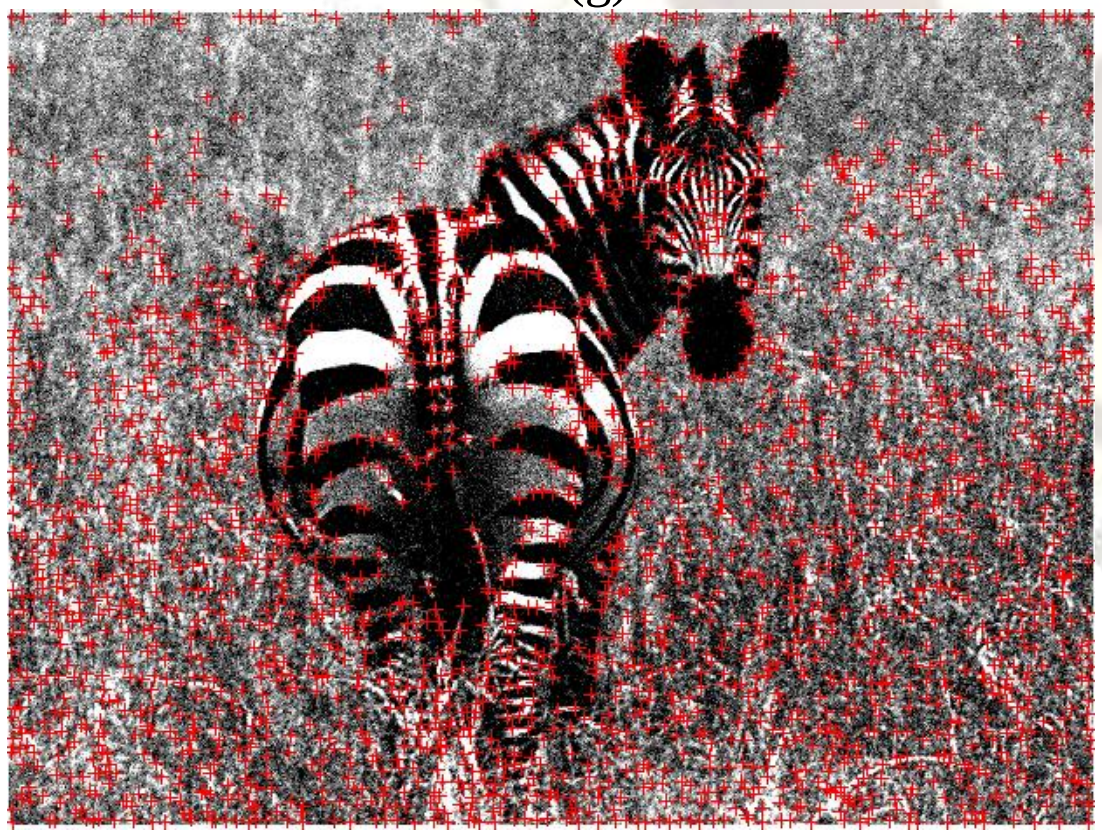

(h)

Table 2

| **Image Type** | | **Moravec Corner detected** | **Harris Corner detected** |
|---|---|---|---|
| Original | | **1722** | **1329** |
| Noise | **Gaussian** (zero mean noise with 0.01 variance) | 2275 | 2259 |
| | **Speckle** (mean 0 and variance v. The default for v is 0.04) | 2023 | 2137 |
| | **Salt & Pepper** (0.05 noise density) | 2471 | 2778 |

Table 3

| Noise Type | Wavelet | Thres-holding | Level of Decom-position | Moravec Corner detected | Harris Corner detected |
|---|---|---|---|---|---|
| Gaussian | Haar | Bayes Soft | 1 | 2035 | 1697 |
| | | | 2 | 1967 | **1379** |
| Speckle | | | 1 | 1962 | 1910 |
| | | | 2 | **1931** | 1623 |
| Salt & Pepper | | | 1 | 2366 | 2624 |
| | | | 2 | 2375 | 2554 |

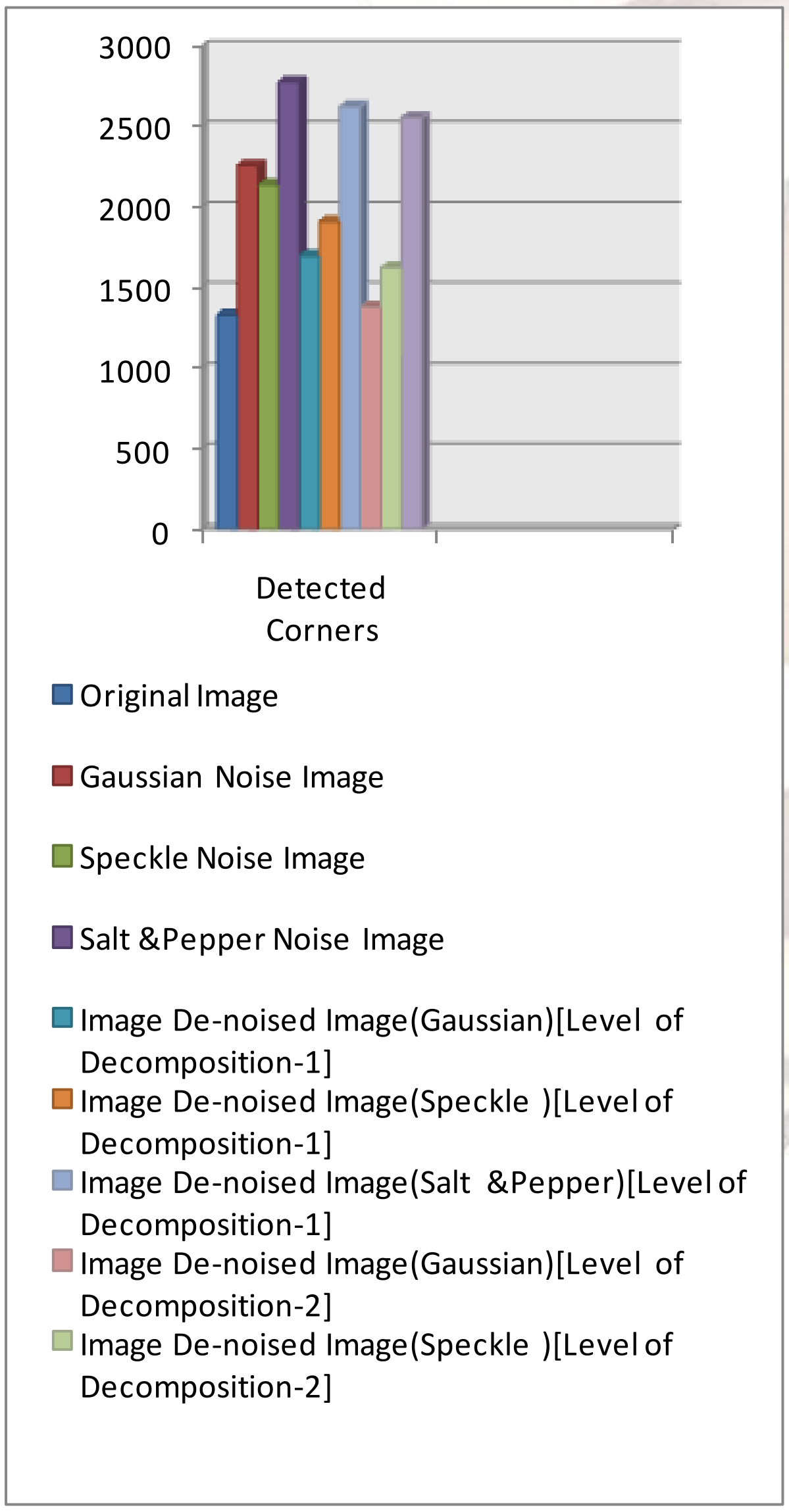

Figure 8. Harris Corner Detection

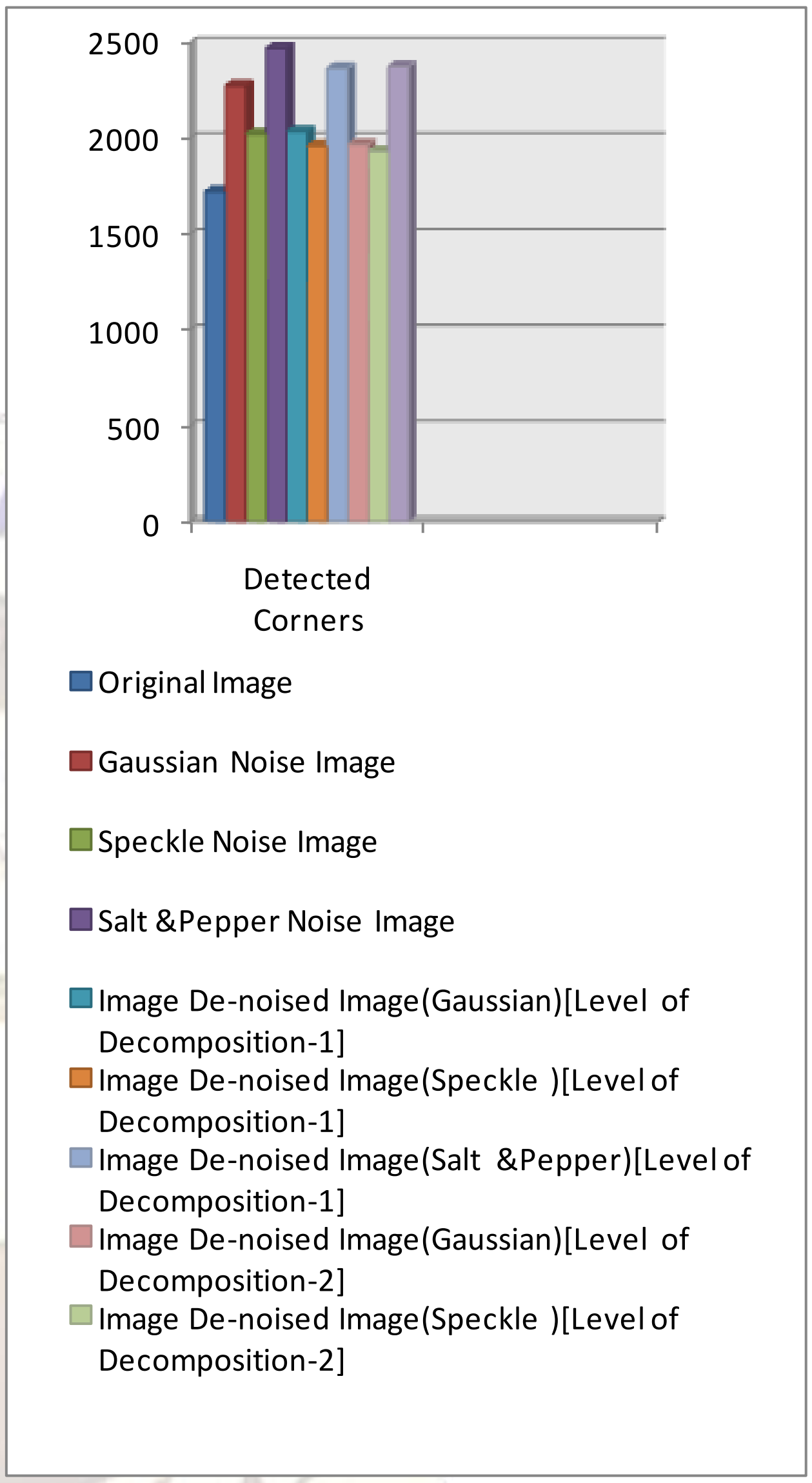

Figure 9. Moravec Corner Detection

## VII. Conclusion

The BS method is effective for images including Gaussian noise. As the experimental result shows that the number of Harris corner detected for obtaining features from the original image is near equal to the same with the number of points detected by de-noised image using BS method.

## IX. References

[1] J. N. Ellinas, T. Mandadelis, A. Tzortzis, L. Aslanoglou, "Image de-noising using wavelets", T.E.I. of Piraeus Applied Research Review, vol. IX, no. 1, pp. 97-109, 2004.

[2] Lakhwinder Kaur and Savita Gupta and R.C.Chauhan, "Image denoising using wavelet thresholding", ICVGIP, Proceeding of the Third Indian Conference On Computer Vision, Graphics & Image Processing, Ahmdabad, India Dec. 16-18, 2002.

[3] Maarten Janse," Noise Reduction by Wavelet Thresholding", Volume 161, Springer Verlag, States United of America, I edition, 2000

[4] D. L. Donoho, "Denoising by soft-thresholding," IEEE Trans. Inf. Theory, vol. 41, no. 3, pp. 613–627, Mar. 1995.

[5] D. L. Donoho, De-Noising by Soft Thresholding, IEEE Trans. Info. Theory 43, pp. 933-936, 1993

[6] D. L. Donoho and I. M. Johnstone, "Ideal spatial adaptation by wavelet shrinkage," Biometrika, vol. 81, no. 3, 1994,pp. 425–455.

[7] J. K. Romberg, H. Choi, R. G. Baraniuk, "Bayesian treestructured image modeling using wavelet-domain hidden Markov models," IEEETrans. Image Process, Vol. 10, No.7, pp. 1056–1068, Jul. 2001.

[8] Harris, C., Stephens, M., 1988, A Combined Corner and Edge Detector, Proceedings of 4th AlveyVision Conference

[9] KonstantinosG. Derpanis, 2004, The Harris Corner Detector

[10] H. P. Moravec. Towards Automatic Visual Obstacle Avoidance. Proc. 5th International Joint Conference on Artificial Intelligence, pp. 584, 1977.

[11] H. P. Moravec. Visual Mapping by a Robot Rover. International Joint Conference on Artificial Intelligence, pp. 598-600, 1979.

[12] http://kiwi.cs.dal.ca/~dparks/CornerDetection/moravec.htm